\documentclass[conference,a4paper]{IEEEtran}
\IEEEoverridecommandlockouts
\usepackage{cite}
\usepackage{amsmath,amssymb,amsfonts}
\usepackage{graphicx}
\usepackage{textcomp}
\usepackage{xcolor}
\usepackage{url}
\usepackage{booktabs}
\usepackage{balance}
\usepackage{placeins}
\usepackage{microtype}
\usepackage[utf8]{inputenc}
\usepackage[hidelinks]{hyperref}
\hypersetup{pdftitle={Shallow Probes Are Sufficient for Near-Oracle Personalization Depth Allocation in Federated Learning},pdfauthor={Beyazit Bestami Yuksel, Emrah Dikbiyik, Ayse Yilmazer Metin},pdfkeywords={federated learning, personalization depth, probe sufficiency},pdfcreator={pdfTeX},pdfproducer={LaTeX with hyperref}}

\renewcommand{\refname}{References}
\newcommand{\fedpba}{\textsc{FedPBA}}
\begin{document}
\title{Shallow Probes Are Sufficient for Near-Oracle\\Personalization Depth Allocation in Federated Learning}

\author{\IEEEauthorblockN{1\textsuperscript{st} Beyaz{\i}t Bestami Y\"{U}KSEL}
\IEEEauthorblockA{\textit{Department of Computer Engineering} \\
\textit{Istanbul Technical University}\\
Istanbul, Turkey \\
yukselbe18@itu.edu.tr}
\and
\IEEEauthorblockN{2\textsuperscript{nd} Emrah D\.{I}KBIYIK}
\IEEEauthorblockA{\textit{Department of Computer Engineering} \\
\textit{Marmara University}\\
Istanbul, Turkey \\
emrah.dikbiyik@marmara.edu.tr}
\and
\IEEEauthorblockN{3\textsuperscript{rd} Ay\c{s}e YILMAZER MET\.{I}N}
\IEEEauthorblockA{\textit{Department of Computer Engineering} \\
\textit{Istanbul Technical University}\\
Istanbul, Turkey \\
yilmazerayse@itu.edu.tr}
}

\maketitle

\begin{abstract}
Rather than proposing a new federated optimizer, we study whether a small client-specific measurement budget can support cost-aware personalization-depth decisions under a utility that trades Macro-F1 against deployment cost. Exhaustive evaluation of six depths defines an offline oracle that is not deployable. We audit the Probe Sufficiency Hypothesis: three shallow probes ($P_0$--$P_2$) may contain most decision-relevant information for approaching that oracle. FedPBA is a secondary learned instantiation for deep-action extrapolation. On CIFAR-10/100 and patient-disjoint PTB-XL (45 sessions; 900 clients; 2700 evaluations; GroupKFold), training-free greedy probing achieves regret 0.0071 and 98.3 percent oracle recovery, beating metadata-only policies and FedPBA (0.0110 / 97.3 percent). Aggregate shallow-best versus oracle correlation is 0.9818 but falls to 0.578 on deep-oracle rows; PTB-XL deep-oracle cases remain a high-regret boundary without a validated escalation gate.
\end{abstract}


\section{Introduction}
Federated learning (FL) trains shared models without pooling raw data~\cite{mcmahan2017fedavg,kairouz2021advances}. Personalized FL (PFL) adapts those models via meta-learning, proximal objectives, or personalized layers~\cite{fallah2020perfedavg,li2021ditto,arivazhagan2019fedper,collins2021fedrep,dinh2020pfedme,tan2022surveypfl}. Most work asks \emph{how} to personalize. We instead ask \emph{how deeply} a client should personalize under an explicit performance--cost utility, and whether that depth can be chosen \emph{without} evaluating every candidate.

Deeper local adaptation can raise Macro-F1 but increases parameters, FLOPs, wall time, communication, and memory. Exhaustive evaluation of $\{P_0,\ldots,P_5\}$ defines an offline \emph{oracle} that is informative yet not deployable: measuring every depth on every client is itself costly. A companion characterization study~\cite{anon_char} mapped that offline landscape; here we study the information-sufficiency and policy-design problem required to approximate it under a constrained measurement budget---probe-derived information, policy-class comparison, deep-action extrapolation, and failure conditions.

After FedAvg, we treat $P_0$--$P_2$ as shallow probes that observe no adaptation, head adaptation, and limited representation adaptation. These three measurements sample the early shape of each client's local utility curve under
\begin{equation}
U_i(a)=\mathrm{Macro\text{-}F1}_i(a)-\lambda\,\mathrm{Cost}_i(a).
\label{eq:utility}
\end{equation}
We introduce the Probe Sufficiency Hypothesis and audit four policy classes. Fixed policies use no client-specific probe information. Context-only learners map metadata to depth without measuring utilities. The training-free greedy probe rule is the direct sufficiency test: after three measurements, deploy the shallow optimum. \fedpba{} (Federated Probe-Based Allocation) is a complementary learned probe allocator that uses the same probes plus context to test deep-action extrapolation to $P_3$--$P_5$ without evaluating those depths. We do not present greedy and \fedpba{} as co-equal proposed algorithms: greedy is the primary empirical test of Probe Sufficiency, and \fedpba{} is a learned instantiation of the probe-first framework.

\textbf{Contributions.} (1)~Formulate cost-aware personalization-depth allocation as approximating an offline oracle with fewer evaluations. (2)~Introduce and audit Probe Sufficiency with fixed, context-only, greedy-probe, and learned probe policies under GroupKFold and session-level inference. (3)~Provide evidence and boundary conditions: greedy regret $0.0071$ / $98.3\%$ recovery; \fedpba{} $0.0110$ / $97.3\%$; aggregate shallow-best vs.\ oracle $r{=}0.9818$ (deep-oracle-only $r{=}0.578$); PTB-XL deep-oracle cases remain the main failure boundary without a validated escalation gate. We propose neither a new FL optimizer nor a cryptographic privacy mechanism~\cite{yuksel2026federated}; any privacy benefit inherits only from FL's raw-data locality.

\section{Related Work}
\textbf{Personalized FL.} FedAvg and heterogeneity-robust variants learn shared models under data locality~\cite{mcmahan2017fedavg,li2020fedprox,karimireddy2020scaffold}. PFL methods design objectives, representations, proximal adaptation, meta-learning, or personalized layers~\cite{fallah2020perfedavg,li2021ditto,dinh2020pfedme,arivazhagan2019fedper,collins2021fedrep,tan2022surveypfl}. The missing question is not how to construct a personalized model, but whether a small client-specific measurement budget can support near-oracle selection among personalization depths under an explicit performance--cost utility.

\textbf{Resource-aware FL.} Client selection, compression, heterogeneous capacity, scheduling, and hardware-aware deployments reduce federation-level cost~\cite{lai2021oort,cho2020tifl,stich2018localsgd,nishio2019client,diao2021heterofl,yuksel2026hardware}. These address \emph{who} participates or \emph{how} to communicate. After global training, we ask \emph{how deeply} each participating client should adapt under a cost-aware utility, and whether that depth can be chosen with a limited probe budget.

\textbf{Federated PEFT.} Recent work personalizes \emph{how} adapters are structured or aggregated~\cite{yang2024feddpa,bai2024flexlora,ghiasvand2025fedtt,lee2026fedp2eft,bian2026fedalt,su2025ailora}. To our knowledge, it does not treat three shallow client-specific utility measurements as the control signal for approximating a cost-aware depth oracle.

\textbf{Budgeted computation and companion work.} Probe-first allocation resembles early-exit / budgeted computation as a systems principle: spend a small measurement budget before committing to a costly action. Here the stopping decision is made at the client-level personalization-depth allocation layer, not at an inference-layer exit. We use this analogy only as conceptual framing. Relative to~\cite{anon_char}, which defined and characterized the offline decision landscape, this paper studies operational probe-first approximation, policy-class comparison, and failure-condition analysis.

\section{Problem and Method}
We distinguish \emph{decision formulation} from \emph{decision policy}. Decision formulation defines the action space, utility, cost model, and offline oracle. Decision policy selects an action using information available before exhaustive evaluation. The companion study primarily studied the former; the present paper studies the latter under a constrained shallow probe budget.

Clients $i\in\{1,\ldots,N\}$ obtain a global model by FedAvg, then select $a\in\mathcal{A}=\{P_0,\ldots,P_5\}$ (Table~\ref{tab:ladder}). Utility follows Eq.~\eqref{eq:utility}. Macro-F1 is used because client-level label imbalance is common under Dirichlet partitions and in medical classification; we did not compare Accuracy or AUROC in this study.

\noindent\textit{Cost model.} For each action we record trainable parameters, FLOPs, local training wall time, communication bytes, and a memory proxy. Raw cost is a weighted sum of magnitude-scaled components with default weights $(0.25,0.20,0.25,0.15,0.15)$, then multiplied by a client hardware multiplier and min--max normalized \emph{within each client across the six actions}. If the within-client range is below $10^{-12}$, all six normalized costs are set to $0$. Min--max combines heterogeneous units into one utility; $\lambda\in\{0.1,0.3,1.0\}$ then sets the Macro-F1--cost exchange rate as an evaluation regime, not a universal physical constant. Client-wide scalar hardware multipliers cancel under within-client normalization, so the present cost model is \emph{not} hardware-aware for action ranking; action-dependent hardware profiling remains future work. A post-hoc sensitivity analysis over six weight profiles and two normalizations yields greedy regret in $[0.0070,0.0094]$ and recovery in $[97.7\%,98.3\%]$ with shallow-oracle mass in $[91.3\%,95.3\%]$.

\begin{table}[!t]
\caption{Personalization ladder (action space).}
\label{tab:ladder}
\centering
\footnotesize
\begin{tabular}{@{}cl@{}}
\toprule
Action & Locally trainable components\\
\midrule
$P_0$ & None (global-only)\\
$P_1$ & Classifier head\\
$P_2$ & Adapter and head\\
$P_3$ & Last encoder block, adapter, head\\
$P_4$ & Last two encoder blocks, adapter, head\\
$P_5$ & Full model\\
\bottomrule
\end{tabular}
\end{table}

The offline oracle $a_i^{\star}=\arg\max_{a\in\mathcal{A}} U_i(a)$ requires all six depths and is an empirical upper bound only. Regret is $r_i(\pi)=U_i(a_i^{\star})-U_i(\pi_i)$. Mean oracle utility over the evaluation set is positive ($0.4147$), so the recovery ratio
\begin{equation}
\mathrm{Recovery}(\pi)=\frac{\sum_i U_i(\pi_i)}{\sum_i U_i(a_i^{\star})}
\label{eq:recovery}
\end{equation}
is well-defined; recovery would be fragile if the oracle aggregate approached zero or became negative.

\textbf{Probe Sufficiency Hypothesis.} Define $\mathbf{s}_i=[U_i(P_0),U_i(P_1),U_i(P_2)]$ and context $\mathbf{c}_i$ (dataset, Dirichlet $\alpha$, $\lambda$, and related metadata). These probes are meaningful because $P_0$ measures the unadapted global model, $P_1$ measures head-only adaptation, and $P_2$ measures limited representation adaptation. For allocation under Eq.~\eqref{eq:utility} in the evaluated setting, $\mathbf{s}_i$ contains most decision-relevant information for approaching the oracle (empirical, not theoretical). We operationalize this via (i)~lower regret than context-only and fixed policies, (ii)~high recovery, and (iii)~association between shallow-best and oracle utility.

Three related but distinct notions should not be conflated: \textbf{aggregate value agreement} ($\max(U_i(P_0),U_i(P_1),U_i(P_2))$ tracks $U_i(a_i^{\star})$, $r{=}0.9818$); \textbf{shallow-action frequency} ($a_i^{\star}\in\{P_0,P_1,P_2\}$ in 2567/2700 cases); and \textbf{deep-regime predictability} (whether shallow signals predict deep gains when $a_i^{\star}\in\{P_3,P_4,P_5\}$, $r{=}0.578$ on 133 rows). Aggregate $r{=}0.9818$ is partly structural because shallow-best equals oracle utility whenever the oracle is shallow. \emph{High aggregate correlation is not equivalent to validated prediction of deep personalization gains.}

\begin{figure*}[!t]
\centering
\includegraphics[width=0.82\textwidth]{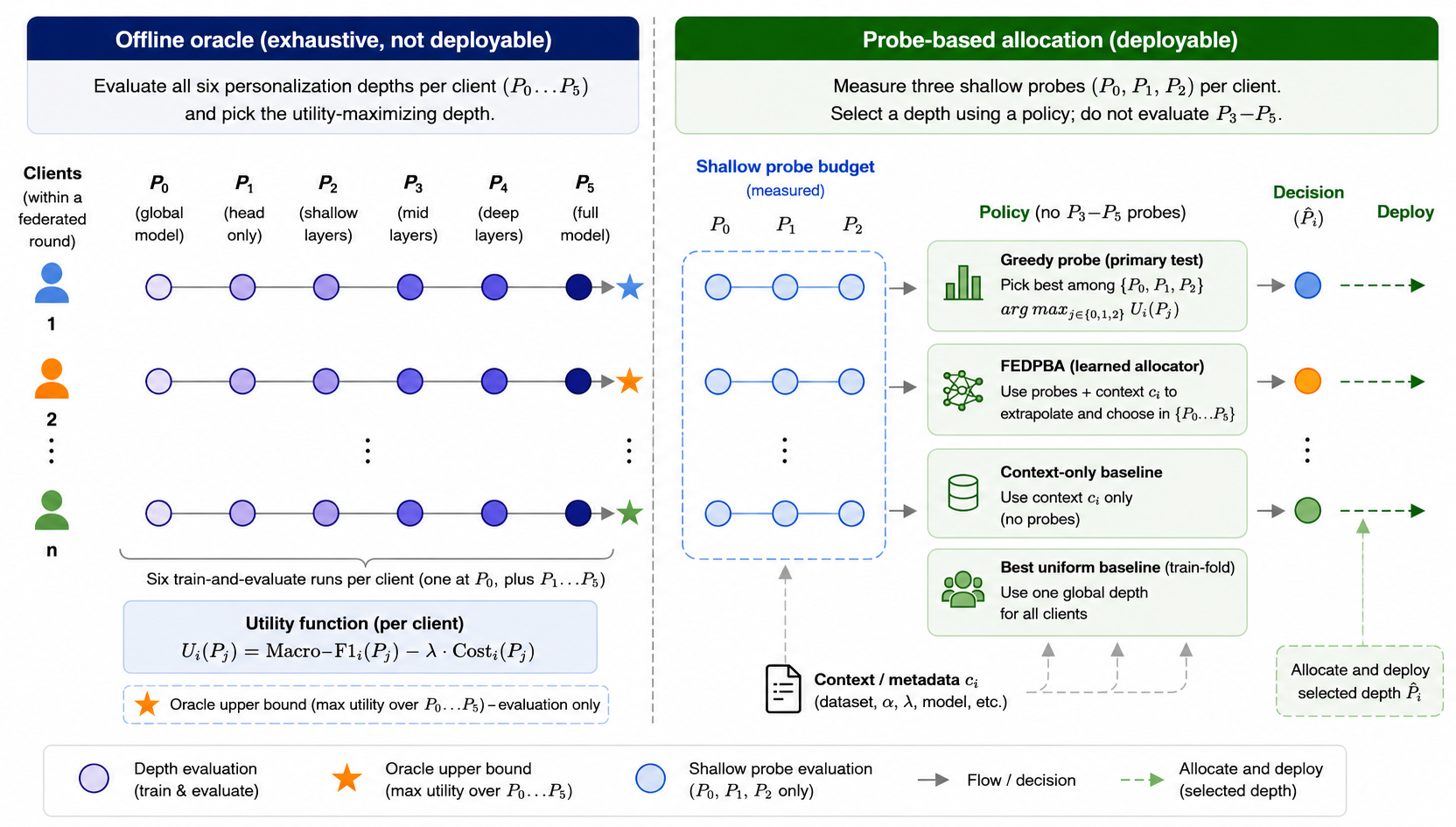}
\caption{Probe-first personalization-depth allocation. Left: offline oracle exhaustively evaluates $P_0$--$P_5$ (empirical upper bound; not deployable). Right: three shallow probes $P_0$--$P_2$ are measured and a policy selects a depth without evaluating $P_3$--$P_5$. Greedy is the direct shallow-sufficiency test; \fedpba{} is the learned extrapolation policy; context-only and best uniform are information-control baselines.}
\label{fig:pipeline}
\end{figure*}

\textbf{Policies.} (A)~Fixed: global-only ($P_0$), personalize-all ($P_5$), majority, and best uniform (one depth per configuration; train-fold only). (B)~Context-only: learners mapping $\mathbf{c}_i\rightarrow a$ without $\mathbf{s}_i$. (C)~Training-free greedy probe: evaluate $P_0$--$P_2$; deploy $\arg\max_{a\in\{P_0,P_1,P_2\}}U_i(a)$---the direct sufficiency test. (D)~\fedpba{}: a meta-level histogram gradient-boosting classifier on probes+context ($\mathtt{max\_iter}{=}150$, $\mathtt{max\_depth}{=}6$, $\mathtt{learning\_rate}{=}0.05$) that may predict $P_3$--$P_5$, testing whether measured shallow evidence supports deep-action extrapolation. \fedpba{} does not alter FedAvg training; it operates after shallow measurements are obtained. Fig.~\ref{fig:pipeline} contrasts the offline oracle with deployable probe-first allocation.

After the three probes, greedy stops and deploys the shallow optimum. \fedpba{} may still recommend a deep action without measuring $P_3$--$P_5$. Histogram boosting is used because features are tabular and mixed-scale, interactions may be nonlinear, the offline policy-training set is moderate, and inference overhead is low; this choice is not a methodological novelty claim. Feature ablations in the broader experimental stack indicated that probe outcomes outweigh representation-geometry descriptors for allocation, so we treat probes as the primary control signal. Primary evaluation uses 5-fold GroupKFold by federated session: all $\lambda$ copies of a client remain in the same fold; preprocessing is fit on training folds only; best uniform and majority are selected on the training fold only. Complementary leave-one-dataset-out folds stress-test \emph{learned} transfer; because greedy is training-free, its leave-one-dataset-out aggregate equals the primary aggregate by construction and is not a cross-dataset generalization claim. Deployment never queries the oracle.

\section{Experimental Setup}
\textbf{Datasets and partitions.} CIFAR-10/100 use a ResNet-style 2D encoder. PTB-XL binary Normal vs.\ Abnormal ECG on Lead~II follows a patient-disjoint protocol ($\sim$5000 records / $\sim$4823 patients; z-score normalized; fixed-length resample)~\cite{wagner2020ptbxl}. $N{=}20$ clients. CIFAR samples are partitioned by label-Dirichlet $\alpha\in\{0.1,0.5,1.0\}$ then split locally 70/15/15. PTB-XL assigns patients first to a global 70/15/15 split (once per seed), then distributes each split's patients across clients via Dirichlet $\alpha$ on patient-level labels. Each patient belongs to exactly one client and one global split; integrity checks confirm zero global and zero cross-client patient overlap. Empty CIFAR clients are rebalanced.

\textbf{Models and training.} Encoder blocks + bottleneck adapter (pool $\rightarrow$ Linear$\rightarrow$ReLU$\rightarrow$Linear$\rightarrow$ReLU) + linear head; $P_1$--$P_5$ unfreeze the components in Table~\ref{tab:ladder}. FedAvg: 5 global rounds, 3 local epochs, batch 32, learning rate $10^{-3}$. Personalization uses the same local epoch/LR settings on the client's train split; Macro-F1 is measured on the client's held-out test split.

\textbf{Cost, scale, and statistics.} Client-wide hardware multipliers scale raw cost before within-client normalization; wall time is measured during local training; communication bytes and memory are proxy estimates. $\lambda\in\{0.1,0.3,1.0\}$ is applied post hoc. Scale: 45 federated sessions, 900 client instances, 2700 client--$\lambda$ evaluations. Descriptive means use the 2700-row pool. Primary inference uses $n{=}45$ session aggregates with paired Wilcoxon tests and Holm correction; main confidence intervals are session-cluster bootstrap ($B{=}10{,}000$). We do not treat $n{=}2700$ client--$\lambda$ Wilcoxon tests as primary inference.

\section{Results}
\subsection{Finding 1: Measurement Dominates Metadata}
\begin{figure}[!t]
\centering
\includegraphics[width=0.90\linewidth]{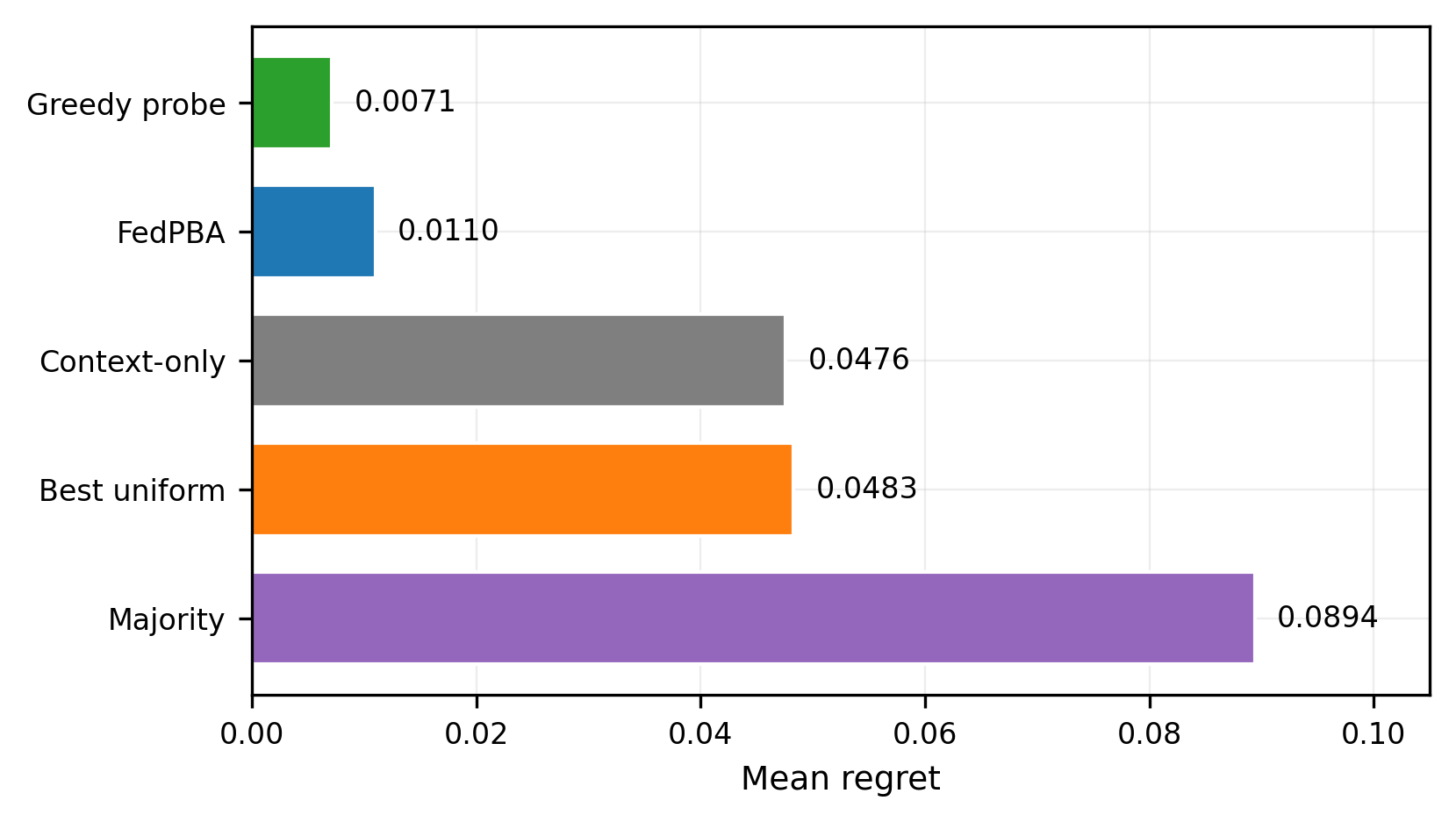}
\caption{Mean regret by policy class ($n{=}2700$). Probe policies sit well below fixed and context-only classes.}
\label{fig:class}
\end{figure}

\begin{figure}[!t]
\centering
\includegraphics[width=0.90\linewidth]{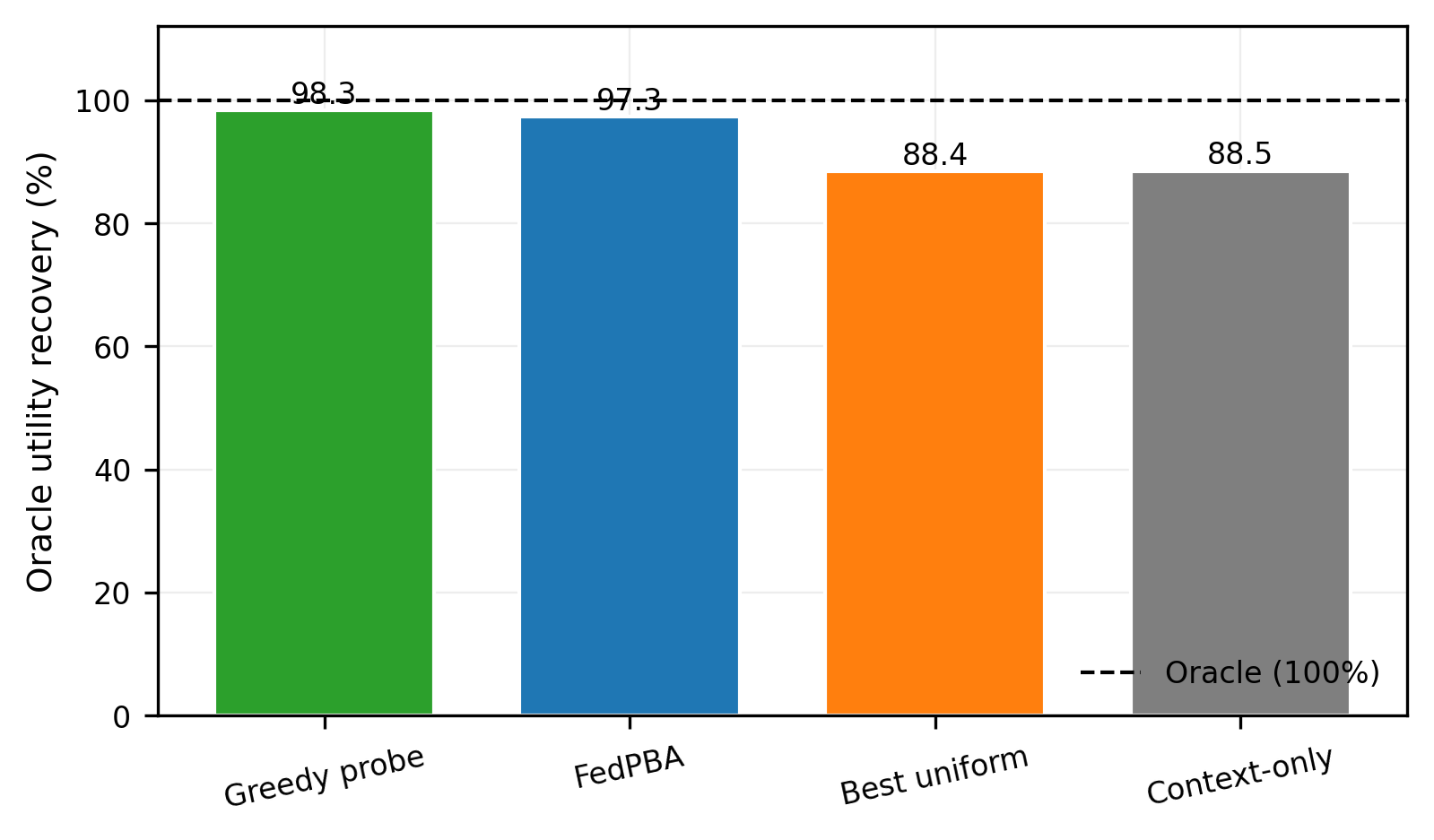}
\caption{Oracle utility recovery across policies (oracle mean $U{=}0.4147$).}
\label{fig:recovery}
\end{figure}

Table~\ref{tab:policy} and Figs.~\ref{fig:class}--\ref{fig:recovery} compare policy classes under the same GroupKFold protocol. Probe policies occupy a distinctly lower-regret / higher-recovery band than fixed and context-only baselines. Context-only tracks best uniform closely ($0.0476$ vs.\ $0.0483$), isolating client-specific utility measurement---not dataset identity or $\alpha$---as the dominant information source. Probe information gain versus context is $0.0405$ (greedy) and $0.0366$ (\fedpba{}); sufficiency ratios versus uniform are $0.852$ and $0.771$. Personalize-all collapses under the cost term (negative aggregate utility), while greedy and \fedpba{} recover nearly all of the positive oracle aggregate. Relative to probe policies, context-only remains far behind; relative to best uniform, the context-only gap is only marginal and is not claimed as a primary paired contrast. Taken together, Figs.~\ref{fig:class}--\ref{fig:recovery} show the same ranking on complementary axes: measurement closes most of the oracle gap, whereas metadata-only allocation does not.

\begin{table}[!t]
\caption{Policy comparison ($n{=}2700$; GroupKFold).}
\label{tab:policy}
\centering
\footnotesize
\setlength{\tabcolsep}{2.5pt}
\begin{tabular}{@{}llccl@{}}
\toprule
Class & Policy & Probes & Regret & Rec.\ \%\\
\midrule
C & Greedy & 3 & $\mathbf{0.0071}$ & $\mathbf{98.3}$\\
D & \fedpba{} & 3+pred. & $0.0110$ & $97.3$\\
A & Best uniform$^{\dagger}$ & 0$^{\ddagger}$ & $0.0483$ & $88.4$\\
B & Best context-only & 0 & $0.0476$ & $88.5$\\
A & Majority & 0 & $0.0894$ & $78.4$\\
A & Personalize-all & 0 & $0.5185$ & $-25.0$\\
\bottomrule
\multicolumn{5}{l}{\footnotesize $^{\dagger}$Train-fold selection. $^{\ddagger}$Uses full-depth outcomes, not probes.}
\end{tabular}
\end{table}

\subsection{Finding 2: Aggregate Agreement Is Not Deep Prediction}
\begin{figure}[!t]
\centering
\includegraphics[width=0.80\linewidth]{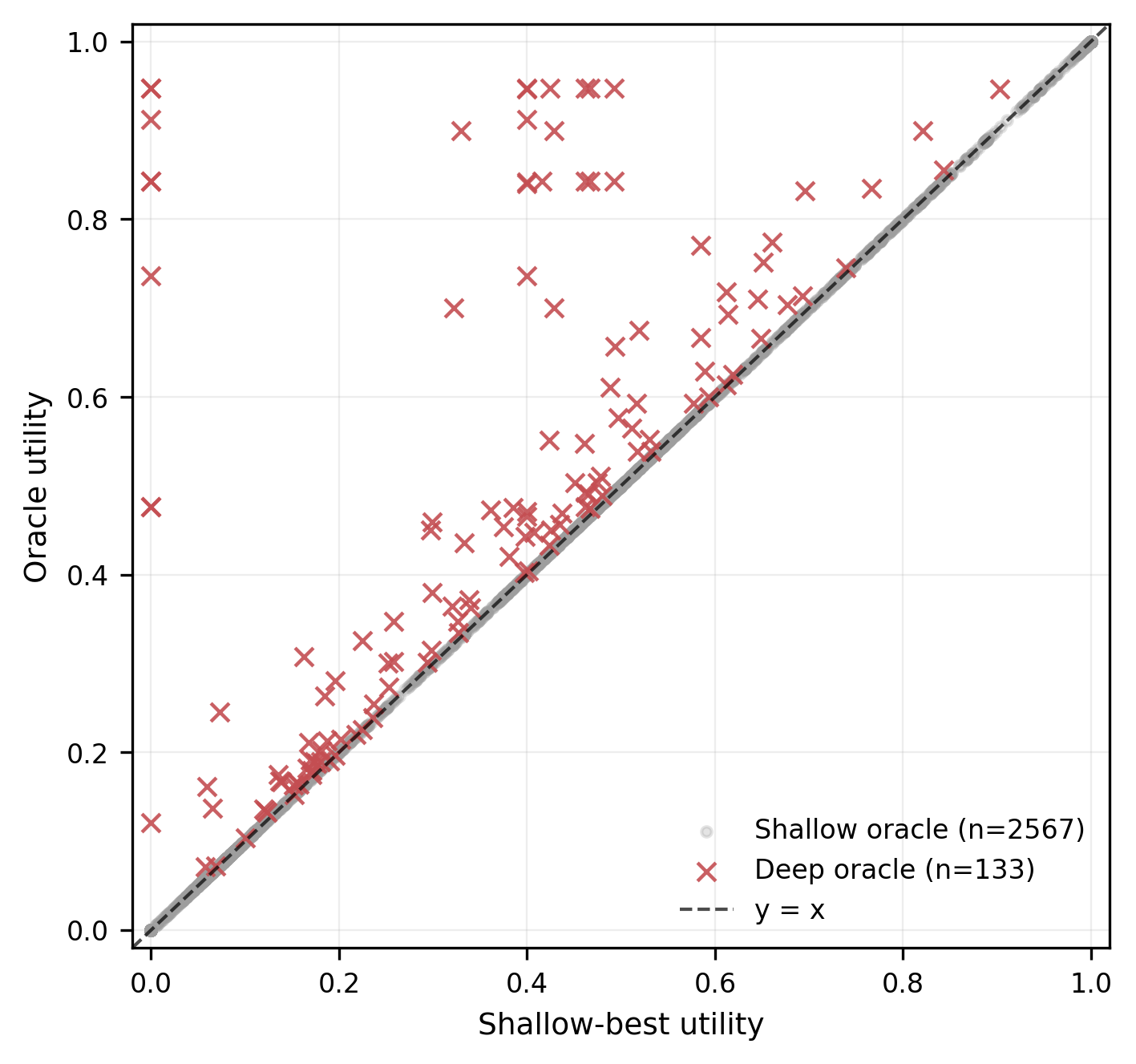}
\caption{Shallow-best vs.\ oracle utility ($r{=}0.9818$ overall, $n{=}2700$; deep-oracle-only $r{=}0.578$, $n{=}133$).}
\label{fig:scatter}
\end{figure}

Fig.~\ref{fig:scatter} and Table~\ref{tab:mechanism} show that $\max(P_0,P_1,P_2)$ tracks oracle utility tightly in aggregate ($r{=}0.9818$), while individual probes remain informative but weaker. On shallow-oracle rows the agreement is structural ($r{=}1$ by construction). Restricting to the 133 deep-oracle rows drops correlation to $0.578$ (95\% CI $[0.28,0.86]$), and early shallow deltas do not positively predict deep gain (e.g., $\Delta_{01}$ vs.\ deep gain $r{=}-0.37$). The oracle is shallow in $95.1\%$ of rows; high aggregate $r$ therefore mixes a large shallow mass with a small, harder deep regime. Reading the scatter as a universal sufficiency proof would overclaim: aggregate recovery can look excellent even when deep-gain prediction remains unresolved.

\begin{table}[!t]
\caption{Probe utility versus oracle utility (Pearson $r$).}
\label{tab:mechanism}
\centering
\footnotesize
\begin{tabular}{@{}lc@{}}
\toprule
Probe signal & $r$\\
\midrule
$P_0$ utility & $0.8353$\\
$P_1$ utility & $0.8605$\\
$P_2$ utility & $0.8323$\\
$\max(P_0,P_1,P_2)$ utility & $\mathbf{0.9818}$\\
\bottomrule
\end{tabular}
\end{table}

\subsection{Finding 3: Probing Beats Allocator Complexity}
\begin{figure}[!t]
\centering
\includegraphics[width=0.82\linewidth]{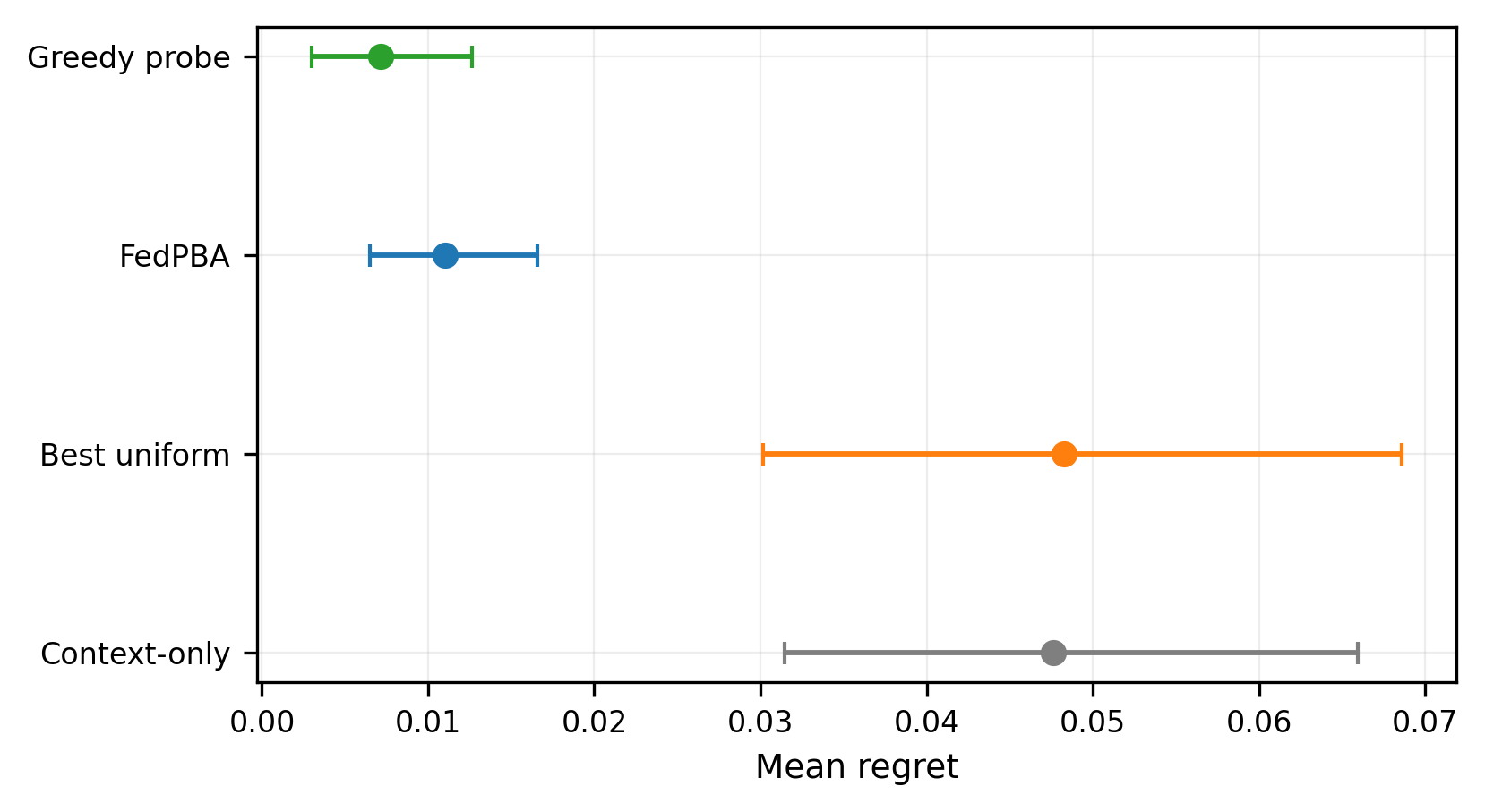}
\caption{Mean regret with session-cluster 95\% bootstrap CI.}
\label{fig:regret}
\end{figure}

The dominant improvement comes from access to client-specific shallow utilities, not from allocator complexity: context-only $\approx$ best uniform, probe policies are substantially better, and training-free greedy outperforms \fedpba{} overall. Session-level tests (Table~\ref{tab:stats}) confirm large gains versus best uniform ($\Delta U{=}+0.0411$ / $+0.0372$; $d_z{=}0.72$ / $0.67$) and a smaller but consistent greedy--\fedpba{} gap ($\Delta U{=}+0.0039$; $d_z{=}1.05$). Bootstrap regret CIs (Fig.~\ref{fig:regret}) do not overlap the best-uniform band. On deep-oracle rows, mean regrets remain similar (greedy $0.145$, \fedpba{} $0.143$; paired utility CI includes zero), so learning does not reliably solve the deep regime.

\begin{table}[!t]
\caption{Session-level paired tests ($n{=}45$; Holm-adjusted).}
\label{tab:stats}
\centering
\scriptsize
\setlength{\tabcolsep}{2pt}
\resizebox{\linewidth}{!}{%
\begin{tabular}{@{}lcccc@{}}
\toprule
Comparison & $\Delta\bar{U}$ & 95\% CI & Holm $p$ & $d_z$\\
\midrule
Greedy vs.\ unif. & $+0.0411$ & $[0.026,0.059]$ & $1.7{\times}10^{-13}$ & $0.72$\\
\fedpba{} vs.\ unif. & $+0.0372$ & $[0.022,0.054]$ & $1.7{\times}10^{-13}$ & $0.67$\\
Greedy vs.\ \fedpba{} & $+0.0039$ & $[0.003,0.005]$ & $1.7{\times}10^{-13}$ & $1.05$\\
\bottomrule
\end{tabular}}
\end{table}

\begin{figure}[!t]
\centering
\includegraphics[width=0.82\linewidth]{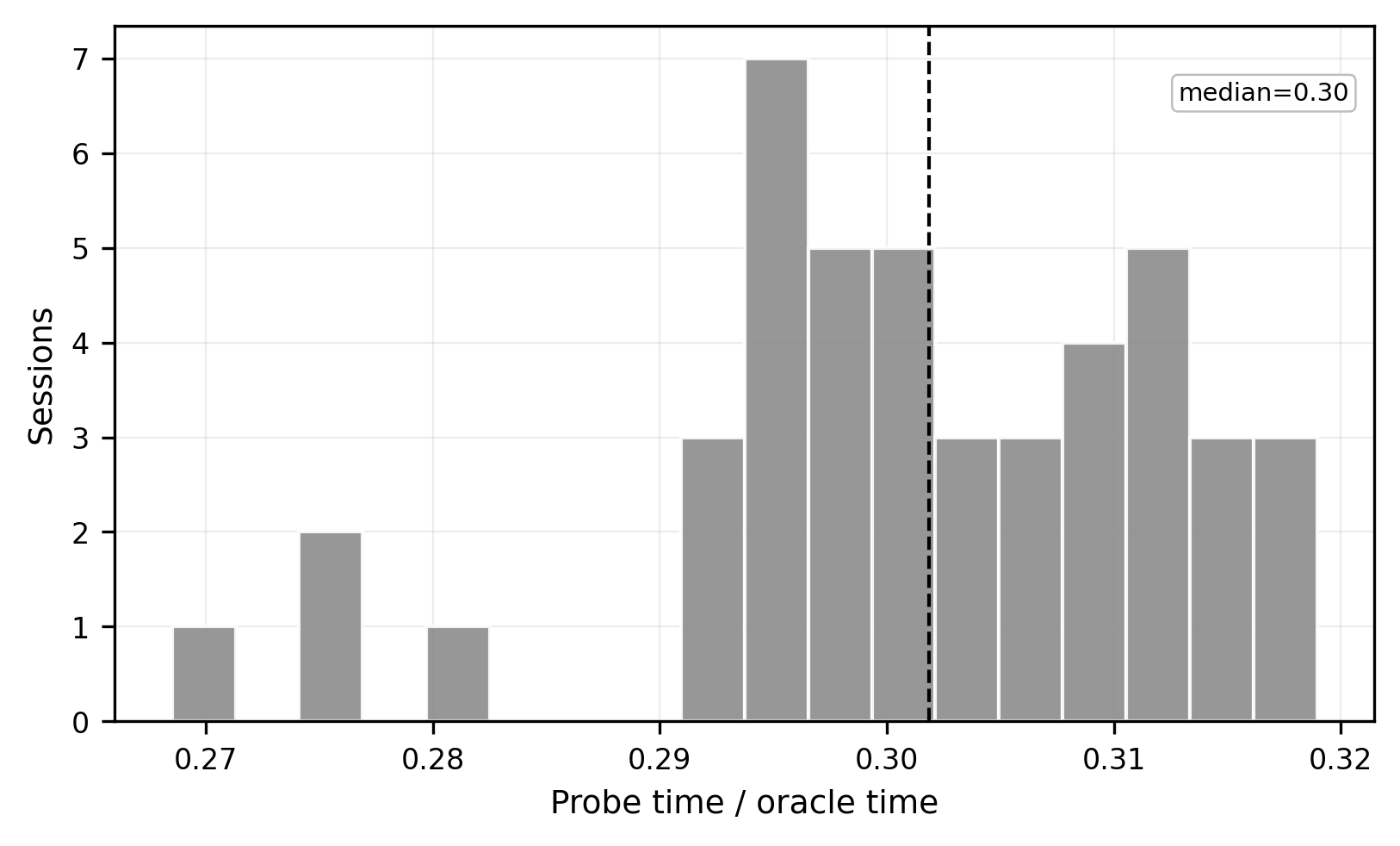}
\caption{Probe-to-exhaustive wall-time ratios across sessions (median ratio $\approx 0.30$).}
\label{fig:speedup}
\end{figure}

Relative to exhaustive search, probing uses 3 vs.\ 6 depth evaluations (and 2 vs.\ 5 personalization train-and-evaluate runs, since $P_0$ is evaluation-only), with a median $\sim$3$\times$ wall-time speedup (Fig.~\ref{fig:speedup}). Leave-one-dataset-out raises \fedpba{} regret to $0.0298$ / $92.8\%$ recovery, exposing calibration dependence; greedy is unchanged. The practical message is therefore not ``prefer the simpler rule because it is simpler,'' but ``prefer the rule whose information source is the three measured utilities and whose performance does not hinge on cross-dataset calibration.''

\subsection{Finding 4: PTB-XL and Deep-Oracle Define the Boundary}
\begin{figure}[!t]
\centering
\includegraphics[width=0.82\linewidth]{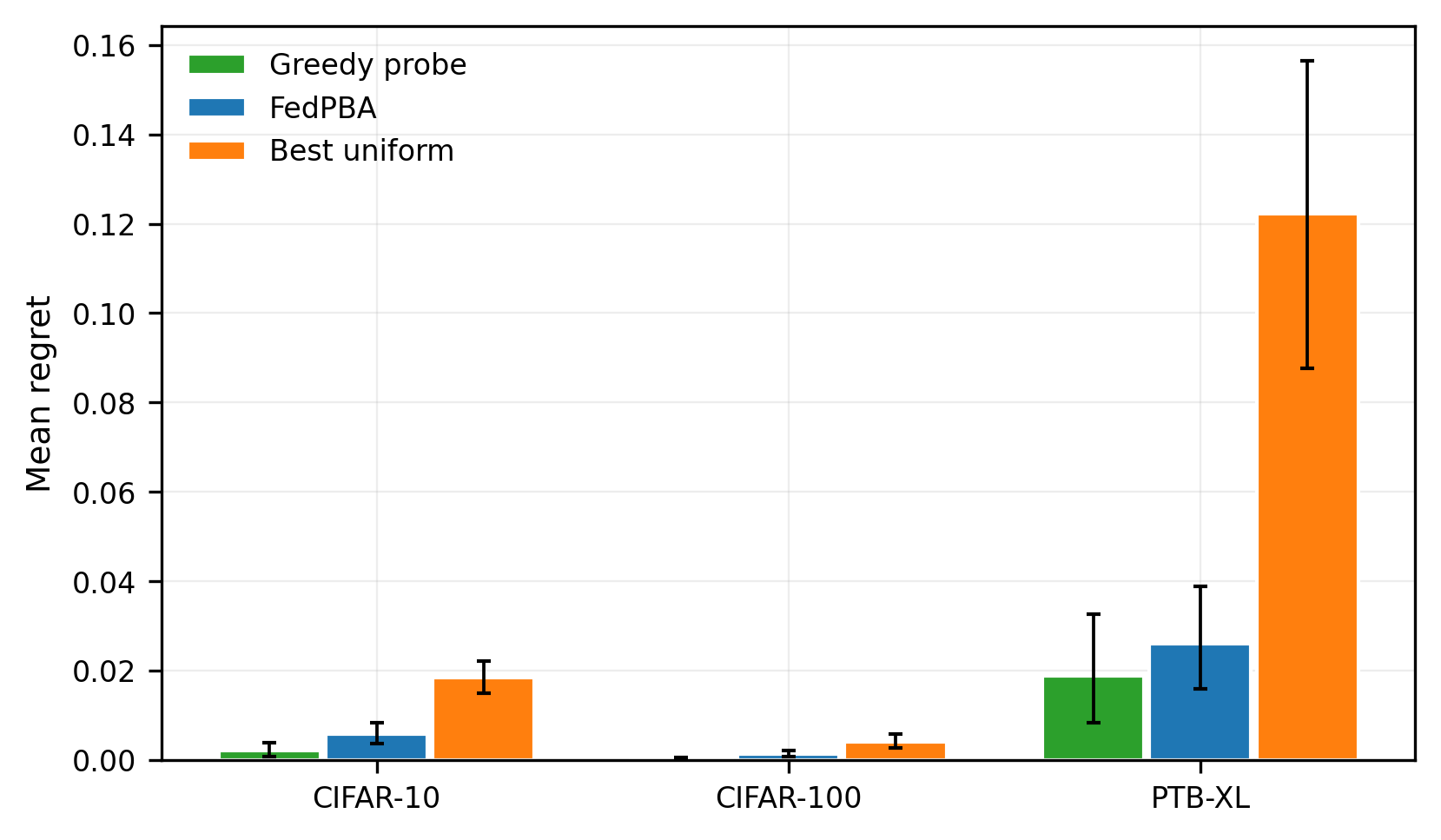}
\caption{Per-dataset mean regret (session-cluster 95\% CI).}
\label{fig:dataset}
\end{figure}

\begin{figure}[!t]
\centering
\includegraphics[width=0.82\linewidth]{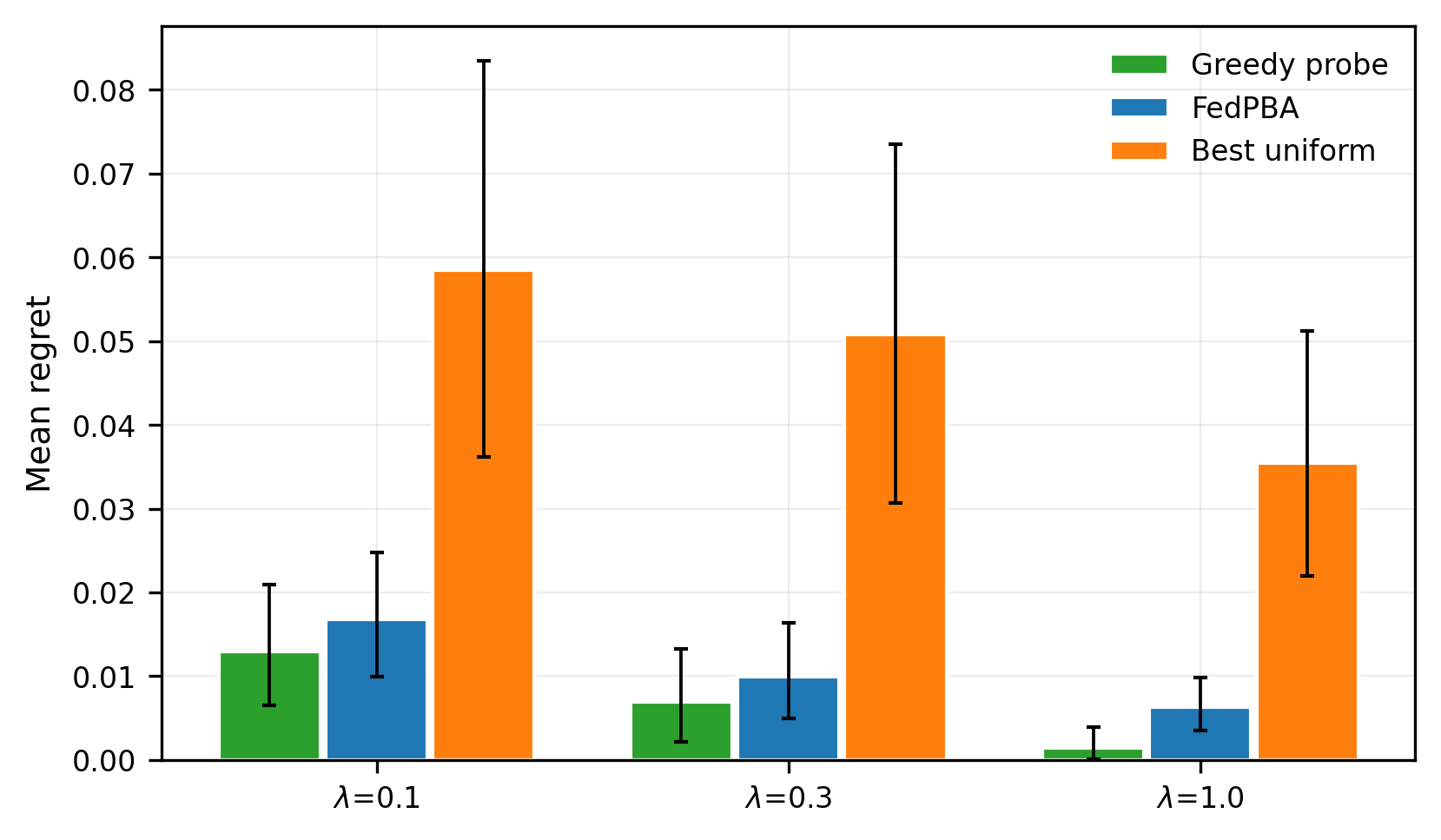}
\caption{Mean regret by cost weight $\lambda$ (session-cluster 95\% CI).}
\label{fig:lambda}
\end{figure}

Fig.~\ref{fig:dataset} separates datasets and Fig.~\ref{fig:lambda} separates $\lambda$ regimes. CIFAR-10/100 stay low-regret under probe policies (CIFAR-10 greedy $0.0022$, \fedpba{} $0.0058$, uniform $0.0184$; CIFAR-100 $0.0003$, $0.0014$, $0.0040$), while absolute regret peaks on patient-disjoint PTB-XL (greedy $0.0189$, \fedpba{} $0.0259$, uniform $0.1223$). The $\lambda$ panels show regret falling as cost weight rises, so low-$\lambda$ regimes leave more room for costly deep mistakes.

\begin{figure}[!t]
\centering
\includegraphics[width=0.82\linewidth]{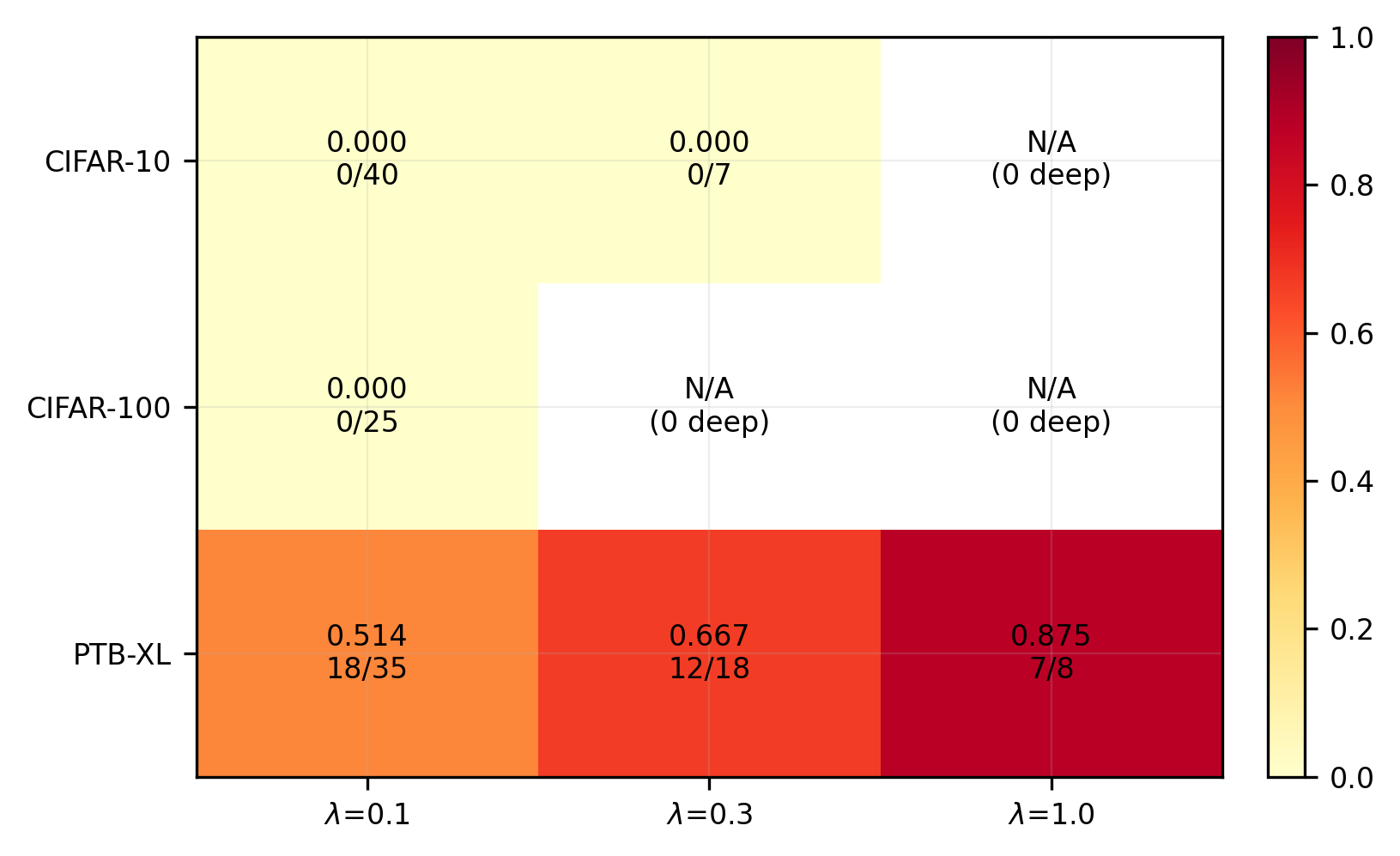}
\caption{Conditional rate $P(\mathrm{greedy}{=}P_0\mid\textrm{oracle deep})$ by dataset and $\lambda$ (N/A if no deep-oracle rows).}
\label{fig:heatmap}
\end{figure}

When the oracle prefers deep actions, greedy has 37 $P_0{\rightarrow}$deep failures (all PTB-XL; mean regret $0.3665$) and \fedpba{} 39 (Table~\ref{tab:failures}). Fig.~\ref{fig:heatmap} localizes $P(\mathrm{greedy}{=}P_0\mid\textrm{oracle deep})$ by dataset and $\lambda$: the mass concentrates on PTB-XL at low-to-moderate $\lambda$, matching the absolute-regret ranking rather than uniform rare-event noise.

\begin{table}[!t]
\caption{$P_0$ selected while oracle prefers $\{P_3,P_4,P_5\}$.}
\label{tab:failures}
\centering
\footnotesize
\begin{tabular}{@{}lcc@{}}
\toprule
 & Greedy & \fedpba{}\\
\midrule
Count & 37 (all PTB-XL) & 39 (38 PTB-XL; 1 CIFAR-10)\\
Mean regret & $0.3665$ & $0.3512$\\
Regret ${>}0.05$ & 34 & 35\\
\bottomrule
\end{tabular}
\end{table}

These failures are rare in the full pool but expensive when they occur. Across the full deep-oracle subset ($n{=}133$), mean regrets remain high for both policies. This concentration aligns with ECG FL heterogeneity and site/patient shift~\cite{rieke2020digital,strodthoff2021ecg,tang2021ecgpfl,lin2022fedcluster,lin2024edgefl,chorney2024ftl,sathi2025xfl}: unconditional greedy is not a validated default for high-risk ECG settings~\cite{yuksel2026federated}. Rising $P_0$--$P_2$ trends and probe ambiguity are candidate escalation indicators only; no gate is trained or validated here.

\section{Discussion, Limitations, and Conclusion}
\textbf{What holds.} Within the evaluated regime, a limited shallow measurement budget recovers near-oracle aggregate utility. Probe-first allocation is therefore a promising deployment-control principle, but not a claim of universal shallow sufficiency. The headline pattern is stable across the tested cost-weight and normalization profiles (greedy regret in $[0.0070,0.0094]$).

\textbf{Why measurement matters.} Client-specific probe utilities contain information that dataset identity, $\alpha$, and $\lambda$ do not capture. Context-only remains close to best uniform and far from probe policies, isolating measurement---not metadata---as the dominant signal. That isolation is the main control contrast of the study: without $\mathbf{s}_i$, even a flexible learner cannot close the gap to the offline oracle.

\textbf{Why learning is not the main result.} The learned allocator does not outperform the training-free rule consistently. Greedy is strongest overall; on deep-oracle rows, \fedpba{} does not reliably outperform it (paired session CI includes zero). Leave-one-dataset-out further increases \fedpba{} regret, indicating calibration dependence. The scientific result is the value of probing, not gradient-boosting complexity.

\textbf{Correlation caveat and failure boundary.} Aggregate $r{=}0.9818$ reflects value agreement and is partly structural on shallow-oracle rows; deep-oracle-only correlation is $r{=}0.578$. Errors concentrate on deep-oracle rows, PTB-XL, and low $\lambda$ (Table~\ref{tab:failures}), consistent with medical FL observations that client shift can dominate aggregate trends~\cite{rieke2020digital,tang2021ecgpfl}. Escalation indicators (rising $P_0$--$P_2$ trends, probe ambiguity) remain unvalidated.

\textbf{Operational guidance.} After FedAvg, evaluate $P_0$--$P_2$; prefer the greedy shallow optimum when $\lambda$ is medium/high and domain risk is low; recalibrate after architecture, population, or cost-definition changes. Relative to~\cite{anon_char}, this paper adds an operational approximation audit rather than another fixed-depth characterization.

\textbf{Limitations.} Findings are conditional on three datasets (vision + one binary ECG subset), ResNet-style backbones, five seeds, one ladder, a limited $\lambda$ grid, and the tested cost profiles. There is no NLP/transformer/LLM experiment. Primary inference rests on 45 sessions; deep-oracle cases number only 133 rows. The offline oracle is an empirical upper bound. Probe-first allocation is not free (three depth evaluations; two train-and-evaluate runs plus $P_0$ evaluation-only). There is no formal regret bound, no cryptographic privacy mechanism~\cite{yuksel2026federated}, and no validated escalation gate. Communication/memory costs are proxies; client-wide hardware multipliers cancel under within-client normalization. Construct validity of ``depth'' is tied to our ladder. Cost-weight and normalization sensitivity analyses support stability across the tested six$\times$two profiles, but do not exhaust all deployment cost definitions. Results do not establish universal shallow sufficiency or immediate clinical deployment without domain-specific validation.

\textbf{Conclusion.} Three shallow probes recover most aggregate oracle utility: greedy achieves $0.0071$ regret and $98.3\%$ recovery, while \fedpba{} is close but not superior ($0.0110$ / $97.3\%$). High-regret deep-oracle PTB-XL failures remain, and no validated escalation detector exists. The contribution is a probe-first deployment-control principle: in this regime, client-specific measurement is more valuable than allocator complexity, but the principle remains risk-conditional and should be recalibrated when the personalization ladder, population, or cost model changes.

\section*{Acknowledgments}
Fig.~\ref{fig:pipeline} used ChatGPT image generation under IEEE AI disclosure; experiments and analyses are the authors'.

\balance


\begin{thebibliography}{00}
\bibitem{mcmahan2017fedavg} H.~B. McMahan, E.~Moore, D.~Ramage, S.~Hampson, and B.~A. y~Arcas, ``Communication-efficient learning of deep networks from decentralized data,'' in \emph{Proc. AISTATS}, 2017, pp. 1273--1282.
\bibitem{kairouz2021advances} P.~Kairouz et~al., ``Advances and open problems in federated learning,'' \emph{Found. Trends Mach. Learn.}, vol.~14, no.~1--2, pp. 1--210, 2021.
\bibitem{fallah2020perfedavg} A.~Fallah, A.~Mokhtari, and A.~Ozdaglar, ``Personalized federated learning with theoretical guarantees: A model-agnostic meta-learning approach,'' \emph{Adv. Neural Inf. Process. Syst.}, vol.~33, 2020.
\bibitem{li2021ditto} T.~Li, S.~Hu, A.~Beirami, and V.~Smith, ``Ditto: fair and robust federated learning through personalization,'' in \emph{Proc. ICML}, 2021, pp. 6357--6368.
\bibitem{arivazhagan2019fedper} M.~G. Arivazhagan, V.~Aggarwal, A.~K. Singh, and S.~Choudhary, ``Federated learning with personalization layers,'' \emph{arXiv:1912.00818}, 2019.
\bibitem{collins2021fedrep} L.~Collins, H.~Hassani, A.~Mokhtari, and S.~Shakkottai, ``Exploiting shared representations for personalized federated learning,'' in \emph{Proc. ICML}, 2021, pp. 2089--2099.
\bibitem{dinh2020pfedme} C.~T. Dinh, N.~Tran, and J.~Nguyen, ``Personalized federated learning with Moreau envelopes,'' \emph{Adv. Neural Inf. Process. Syst.}, vol.~33, 2020.
\bibitem{tan2022surveypfl} A.~Z. Tan, H.~Yu, L.~Cui, and Q.~Yang, ``Towards personalized federated learning,'' \emph{IEEE Trans. Neural Netw. Learn. Syst.}, vol.~34, no.~12, pp. 9587--9603, 2023.
\bibitem{wagner2020ptbxl} P.~Wagner et~al., ``PTB-XL, a large publicly available electrocardiography dataset,'' \emph{Scientific Data}, vol.~7, no.~1, p.~154, 2020.
\bibitem{li2020fedprox} T.~Li, A.~K. Sahu, M.~Zaheer, M.~Sanjabi, A.~Talwalkar, and V.~Smith, ``Federated optimization in heterogeneous networks,'' \emph{Proc. Mach. Learn. Syst.}, vol.~2, pp. 429--450, 2020.
\bibitem{karimireddy2020scaffold} S.~P. Karimireddy et~al., ``SCAFFOLD: stochastic controlled averaging for federated learning,'' in \emph{Proc. ICML}, 2020, pp. 5132--5143.
\bibitem{lai2021oort} F.~Lai, X.~Zhu, H.~V. Madhyastha, and M.~Chowdhury, ``Oort: efficient federated learning via guided participant selection,'' in \emph{Proc. USENIX OSDI}, 2021, pp. 19--35.
\bibitem{cho2020tifl} Z.~Chai et~al., ``TiFL: a tier-based federated learning system,'' in \emph{Proc. ACM HPDC}, 2020, pp. 125--136.
\bibitem{stich2018localsgd} S.~U. Stich, ``Local SGD converges fast and communicates little,'' in \emph{Proc. ICLR}, 2019.
\bibitem{nishio2019client} T.~Nishio and R.~Yonetani, ``Client selection for federated learning with heterogeneous resources in mobile edge,'' in \emph{Proc. IEEE ICC}, 2019, pp. 1--7.
\bibitem{diao2021heterofl} E.~Diao, J.~Ding, and V.~Tarokh, ``HeteroFL: computation and communication efficient federated learning for heterogeneous clients,'' in \emph{Proc. ICLR}, 2021.
\bibitem{yang2024feddpa} Y.~Yang, G.~Long, T.~Shen, J.~Jiang, and M.~Blumenstein, ``Dual-personalizing adapter for federated foundation models,'' \emph{Adv. Neural Inf. Process. Syst.}, vol.~37, 2024.
\bibitem{bai2024flexlora} J.~Bai, D.~Chen, B.~Qian, L.~Yao, and Y.~Li, ``Federated fine-tuning of large language models under heterogeneous tasks and client resources,'' \emph{Adv. Neural Inf. Process. Syst.}, vol.~37, 2024.
\bibitem{ghiasvand2025fedtt} S.~Ghiasvand et~al., ``Communication-efficient and tensorized federated fine-tuning of large language models,'' in \emph{Findings of ACL}, 2025, pp. 24192--24207.
\bibitem{lee2026fedp2eft} R.~Lee, M.~Kim, F.~Rezk, R.~Li, S.~I. Venieris, and T.~Hospedales, ``FedP$^2$EFT: federated learning to personalize PEFT for multilingual LLMs,'' in \emph{Proc. AAAI}, 2026, pp. 22805--22813.
\bibitem{bian2026fedalt} J.~Bian, L.~Wang, L.~Zhang, and J.~Xu, ``FedALT: federated fine-tuning through adaptive local training with rest-of-world LoRA,'' in \emph{Proc. AAAI}, 2026, pp. 19728--19736.
\bibitem{su2025ailora} Y.~Su, N.~Yan, and Y.~Deng, ``Federated LLMs fine-tuned with adaptive importance-aware LoRA,'' in \emph{Proc. IEEE ICC}, 2025, pp. 6112--6117.
\bibitem{anon_char} B.~B. Y\"uksel and A.~Yılmazer Metin, ``When is personalized federated learning worth its deployment cost?'' in \emph{Proc. UBMK}, 2026, to be published.
\bibitem{yuksel2026hardware} B.~B. Y\"uksel and E.~Dikbiyik, ``Hardware-aware federated learning for speech emotion recognition,'' in \emph{Proc. SIU}, 2026, pp. 1--4.
\bibitem{yuksel2026federated} B.~B. Yuksel and A.~Y.~Metin, ``Federated learning with homomorphic encryption for secure real-time ECG anomaly detection,'' \emph{Biomed. Signal Process. Control}, vol.~116, p.~109557, 2026.
\bibitem{rieke2020digital} N.~Rieke et~al., ``The future of digital health with federated learning,'' \emph{npj Digit. Med.}, vol.~3, p.~119, 2020.
\bibitem{tang2021ecgpfl} R.~Tang, J.~Luo, J.~Qian, and J.~Jin, ``Personalized federated learning for ECG classification based on feature alignment,'' \emph{Security and Communication Networks}, vol.~2021, Art.\ no.~6217601, 2021.
\bibitem{strodthoff2021ecg} N.~Strodthoff, P.~Wagner, T.~Schaeffter, and W.~Samek, ``Deep learning for ECG analysis: benchmarks and insights from PTB-XL,'' \emph{IEEE J. Biomed. Health Inform.}, vol.~25, no.~5, pp. 1519--1528, 2021.
\bibitem{lin2022fedcluster} D.~Lin, Y.~Guo, H.~Sun, and Y.~Chen, ``FedCluster: a federated learning framework for cross-device private ECG classification,'' in \emph{Proc. IEEE INFOCOM Workshops}, 2022, pp. 1--6.
\bibitem{lin2024edgefl} H.-T. Lin and C.-Y. Wen, ``Edge federated optimization for heterogeneous data,'' \emph{Future Internet}, vol.~16, no.~4, p.~142, 2024.
\bibitem{chorney2024ftl} W.~Chorney and H.~Wang, ``Towards federated transfer learning in electrocardiogram signal analysis,'' \emph{Comput. Biol. Med.}, vol.~170, p.~107984, 2024.
\bibitem{sathi2025xfl} T.~A. Sathi et~al., ``Explainable federated learning for multi-class heart disease diagnosis via ECG fiducial features,'' \emph{Diagnostics}, vol.~15, no.~24, p.~3110, 2025.
\end{thebibliography}
\end{document}